%% file: main.tex
\documentclass[11pt,letter]{article}
\usepackage[left=1.25in,top=1.0in,right=1.25in,bottom=1.0in]{geometry}
\usepackage{natbib}

\usepackage[utf8]{inputenc} 
\usepackage[T1]{fontenc}    
\usepackage{hyperref}       
\usepackage{url}            
\usepackage{booktabs}       
\usepackage{algorithmic}
\usepackage{amsmath}
\usepackage{amssymb}
\usepackage{amsfonts}       
\usepackage{nicefrac}       
\usepackage{microtype}      
\usepackage[ruled,vlined]{algorithm2e}
\usepackage{xcolor}
\usepackage{cancel}
\usepackage{multicol}
\usepackage[titletoc]{appendix}

\usepackage{graphicx}
\usepackage{caption}
\usepackage{subcaption}

\newcommand{\polya}{Pólya}
\newcommand{\nystrom}{Nystr\"{o}m}

\renewcommand{\b}{\mathbf{b}}
\newcommand{\B}{\mathbf{B}}

\newcommand{\betadist}{\text{Beta}}
\newcommand{\bbeta}{\boldsymbol{\beta}}
\newcommand{\bkappa}{\boldsymbol{\kappa}}

\newcommand{\bmu}{\boldsymbol{\mu}}

\newcommand{\bomega}{\boldsymbol{\omega}}
\newcommand{\bOmega}{\boldsymbol{\Omega}}
\newcommand{\bPhi}{\varphi(\X)}
\newcommand{\bPsi}{\boldsymbol{\Psi}}
\newcommand{\bpsi}{\boldsymbol{\psi}}
\newcommand{\bSigma}{\boldsymbol{\Sigma}}
\newcommand{\btheta}{\boldsymbol{\theta}}
\newcommand{\bvarepsilon}{\boldsymbol{\varepsilon}}
\newcommand{\E}{\mathbb{E}}
\newcommand{\gammadist}{\mbox{Ga}}
\newcommand{\GP}{\mathcal{GP}}

\newcommand{\N}{\mathcal{N}}
\newcommand{\NB}{\text{NB}}
\newcommand{\poisson}{\text{Poisson}}
\newcommand{\rest}{\text{\textemdash}}
\newcommand{\Reals}{\mathbb{R}}
\newcommand{\eye}{\mathbf{I}}
\renewcommand{\d}{\text{d}}

\newcommand{\F}{\mathbf{F}}
\newcommand{\K}{\mathbf{K}}
\newcommand{\m}{\mathbf{m}}

\renewcommand{\S}{\mathbf{S}}
\newcommand{\V}{\mathbf{V}}
\newcommand{\w}{\mathbf{w}}
\newcommand{\W}{\mathbf{W}}
\newcommand{\x}{\mathbf{x}}
\newcommand{\xp}{\x^{\prime}}
\newcommand{\X}{\mathbf{X}}
\newcommand{\y}{\mathbf{y}}
\newcommand{\Y}{\mathbf{Y}}
\newcommand{\z}{\mathbf{z}}

\newcommand{\zero}{\mathbf{0}}

\newcommand{\Ell}{\mathcal{L}}

\title{Latent Variable Modeling with Random Features}
\author{%
    Gregory W. Gundersen\thanks{These authors contributed equally.}
    \quad
    Michael Minyi Zhang\footnotemark[1]
    \quad
    Barbara E. Engelhardt
    \\
    \\
    \textit{Princeton University}
}
\date{June 2020}

\expandafter\def\expandafter\normalsize\expandafter{%
    \normalsize
    \setlength\abovedisplayskip{10pt}
    \setlength\belowdisplayskip{10pt}
    \setlength\abovedisplayshortskip{10pt}
    \setlength\belowdisplayshortskip{10pt}
}

\begin{document}

\maketitle

\input{0_abstract}

\input{1_intro}
\input{2_method}

\input{3_experiments}
\input{4_conclusion}
\clearpage 
\input{0_acknowledgements}


\bibliographystyle{apalike}
\bibliography{references}

\newpage
\input{6_appendix}

\end{document}

%% file: 0_abstract.tex
\begin{abstract}
    Gaussian process-based latent variable models are flexible and theoretically grounded tools for nonlinear dimension reduction, but generalizing to non-Gaussian data likelihoods within this nonlinear framework is statistically challenging.
    Here, we use random features to develop a family of nonlinear dimension reduction models that are easily extensible to non-Gaussian data likelihoods; we call these \emph{random feature latent variable models} (RFLVMs). By approximating a nonlinear relationship between the latent space and the observations with a function that is linear with respect to random features, we induce closed-form gradients of the posterior distribution with respect to the latent variable. This allows the RFLVM framework to support computationally tractable nonlinear latent variable models for a variety of data likelihoods in the exponential family without specialized derivations. Our generalized RFLVMs produce results comparable with other state-of-the-art dimension reduction methods on diverse types of data, including neural spike train recordings, images, and text data.
\end{abstract}

%% file: 1_intro.tex
\section{Introduction}\label{sec:intro}
Many dimension reduction techniques, such as principal component analysis \citep{pearson1901liii, tipping1999probabilistic} and factor analysis \citep{lawley1962factor}, make two modeling assumptions: (1) the observations are Gaussian distributed, and (2) the latent structure is a linear function of the observations. However, for many applications, proper analysis requires us to break both of these assumptions. For example, in computational neuroscience, scientists collect firing rates for thousands of neurons simultaneously. These data are observed as counts, and neuroscientists believe that the biologically relevant latent structure is nonlinear with respect to the observations \citep{saxena2019towards}. 

To capture nonlinear relationships in latent variable models, one approach is to assume that the mapping between the latent manifold and observations is Gaussian process (GP)-distributed. A GP is a prior distribution over the space of real-valued functions, 
which makes posterior inference in GP-based models tractable when the GP prior is conjugate to the data likelihood. This leads to the Gaussian process latent variable model \citep[GPLVM,][]{lawrence2004gaussian}.

The basic GPLVM model with a radial basis function (RBF) kernel has nice statistical properties that allow for exact, computationally tractable inference methods to be used when the number of observations is a reasonable size. 
Deviating from this basic model, however, leads to challenges with inference. 
For Poisson GPLVMs, we cannot integrate out the GP-distributed functional map, and we no longer have closed form expressions for the gradient of the posterior with respect to the latent space. This renders maximum a posteriori (MAP) estimation difficult, leading to solutions at poor local optima \citep[see][]{wu2017gaussian}. 

Random Fourier features \citep[RFFs,][]{rahimi2008random} were developed to avoid working with $N \times N$ dimensional matrices when fitting kernel machines. RFFs accelerate kernel machines by using a low-dimensional, randomized approximation of the inner product associated with a given shift-invariant kernel. For this approximation, RFFs induce a nonlinear map using a linear function of random features. 

We propose to use RFFs to approximate the kernel function in a GPLVM to create a flexible, tractable, and modular framework for fitting GP-based latent variable models. 
In the context of GPLVMs, RFF approximations allow for closed-form gradients of the objective function with respect to the latent variable.
In addition, we can tractably explore the space of stationary covariance functions
by using a Dirichlet process mixture prior for the spectral distribution of frequencies \citep[BaNK,][]{oliva2016bayesian}, leading to a flexible latent variable model. 

This paper makes the following contributions to the space of nonlinear latent variable models: (1) we represent the nonlinear mapping in GPLVMs using a linear function of random Fourier features; (2) we leverage this representation to generalize GPLVMs to non-Gaussian likelihoods; (3) we place a prior on the random features to allow data-driven exploration over the space of shift-invariant kernels, to avoid putting restrictions on the kernel's functional form. 
We validate our approach on diverse simulated data sets, 
and we show how results from RFLVMs compare with state-of-the-art methods on a variety of image, text, and scientific data sets. We release a Python library\footnote{\url{https://github.com/gwgundersen/rflvm}} with modular and extensible code for reproducing and building on our work.

%% file: 2_method.tex
\section{Random Feature Latent Variable Models}
\subsection{Random features for kernel machines}\label{sec:method}
Here we briefly review random Fourier features~\citep{rahimi2008random} to motivate a randomized approximation of the GP-distributed maps in GPLVMs. Bochner's theorem \citep{bochner1959lectures} states that any continuous shift-invariant kernel $k(\x, \xp) = k(\x - \xp)$ on $\Reals^D$ is positive definite if and only if $k(\x - \xp)$ is the Fourier transform of a non-negative measure $p(\w)$. If the kernel is properly scaled, the kernel's Fourier transform $p(\w)$ is guaranteed to be a density. Let $h(\x) \triangleq \exp(i \w^{\top} \x)$, and let $h(\x)^{*}$ denote its complex conjugate. Observe that
\begin{equation}
k(\x - \xp)
=
\int_{\Reals^D} p(\w) \exp(i \w^{\top} (\x - \xp)) \d\w
=
\E_{p(\w)}[h(\x)h(\x)^{*}].
\label{eq:rffs_mc_int}
\end{equation}
So $h(\x)h(\x)^{*}$ is an unbiased estimate of $k(\x - \xp)$. If we drop the imaginary part for real-valued kernels, we can re-define $\smash{h(\x) \triangleq \cos(\w^{\top}\x)}$ by Euler's formula. Then we can use Monte Carlo integration to approximate Eq. \ref{eq:rffs_mc_int} as $k(\x, \xp) \approx \varphi(\x)^{\top} \varphi(\x)$, where
\begin{equation}
    \varphi(\x) \triangleq \sqrt{\frac{2}{M}} \begin{bmatrix}
    \sin(\w_1^{\top} \x) & \cos(\w_1^{\top} \x) & \dots & \sin(\w_{M/2}^{\top} \x) & \cos(\w_{M/2}^{\top} \x)
    \end{bmatrix}^{\top}
\label{eq:rff_def}.
\end{equation}
We draw $M/2$ samples from $p(\w)$, and the definition in Eq.~\ref{eq:rff_def} doubles the number of RFFs to $M$. A representer theorem~\citep{kimeldorf1971some,scholkopf2001generalized} says that the optimal solution to the objective function of a kernel method, $f^{*}(\x)$, is linear in pairwise evaluations of the kernel. Using this random projection, we can represent $f^{*}(\x)$ as
\begin{equation}
f^{*}(\x) 
= \sum_{n=1}^{N} \alpha_n k(\x_n, \x)
= \sum_{n=1}^{N} \alpha_n \langle \phi(\x_n), \phi(\x) \rangle_{\mathcal{H}}
\approx \sum_{n=1}^{N} \alpha_n \varphi(\x_n)^{\top} \varphi(\x)
= \bbeta^{\top} \varphi(\x).
\label{eq:kernel_approx}
\end{equation}
%
In the second equality, the kernel trick implicitly lifts the data into a reproducing kernel Hilbert space $\mathcal{H}$ in which the optimal solution is linear with respect to the features.
The randomized approximation of this inner product allows us to replace expensive calculations involving the kernel with an $M$-dimensional inner product.

For example, the predictive mean in GP regression implicitly uses the representer theorem and kernel trick~\citep{williams2006gaussian}. 
RFFs have been used to reduce the computational costs of fitting GP regression models from $\mathcal{O}(N^3)$ to $\mathcal{O}(NM^2)$~\citep{lazaro2010sparse,hensman2017variational}. However, RFFs have not yet been used to make GPLVMs more computationally tractable.




\subsection{Gaussian process latent variable models}

Now we introduce the basic GPLVM framework~\citep{lawrence2004gaussian}.
Let $\Y$ be an $N \times J$ matrix of $N$ observations and $J$ features, and let $\X$ be an $N \times D$ matrix of latent variables where $D \ll J$. If we take the mean function to be zero, and the observations $\Y$ to be Gaussian distributed, the GPLVM is:
\begin{equation}
\y_j \sim \N_N(f_{j}(\X), \sigma^{2}_j \eye),
\quad
f_{j} \sim \GP(\zero, \K_X),
\quad
\x_n \sim \N_{D}(\zero, \eye),
\label{eq:gplvm_def}
\end{equation}
where $\K_X$ is an $N \times N$ covariance matrix defined by a positive definite kernel function $k(\x, \xp)$ and $f_j(\X) = [f_{j}(\x_1) \dots f_{j}(\x_N)]^{\top}$. Due to conjugacy between the GP prior on $f_{j}$ and Gaussian likelihood on $\y_j$, we can integrate out $f_j$ in closed form. The resulting marginal likelihood for $\y_j$ is $\smash{\N_{N}(0, \K_{X} + \sigma^{2}_{j} \eye)}$. 
We cannot find the optimal $\X$ analytically, but various approximations have been proposed. We can obtain a MAP estimate by integrating out the GP-distributed maps and then optimizing $\X$ with respect to the posterior using scaled conjugate gradients~\citep{lawrence2004gaussian,lawrence2005probabilistic}, where computation scales as $\mathcal{O}(N^3)$. To scale up GPLVM inference, we may use sparse inducing point methods where the computational complexity is $\mathcal{O}(NC^2)$, for $C \ll N$ inducing points \citep{lawrence2007learning}.

Alternatively, we can introduce a variational Bayes approximation of the posterior and minimize the Kullback--Leibler divergence between the posterior and the variational approximation with the latent variables $\X$ marginalized out. However, integrating out $\X$ in the approximate marginal likelihood is only tractable when we assume that we have Gaussian observations and when we use an RBF kernel with automatic relevance determination, which limits its flexibility. This variational approach may be scaled using sparse inducing point methods. This approach is referred to as a \textit{Bayesian GPLVM}~\citep{titsias2010bayesian, damianou2016variational}. 

\subsection{Generative model for RFLVMs}
The generative model of an RFLVM takes the form:
\begin{equation}
\begin{aligned}
\y_j &\sim \Ell \big( g\big( \bPhi \bbeta_j \big), \btheta \big)
&
\w_m &\sim \N_{D}(\bmu_{z_m}, \bSigma_{z_m})
\\
\btheta &\sim p(\btheta)
&
(\bmu_{k}, \bSigma_{k}) &\sim \text{NIW}(\bmu_0, \nu_0, \lambda_0, \bPsi_0)
\\
\bbeta_j &\sim \N_{M}(\bbeta_0, \B_0)
&
z_m &\sim \mbox{CRP}(\alpha),
\\
\quad \x_n &\sim \N_{D}(\zero, \eye)
&
\alpha &\sim \gammadist(a_\alpha, b_\alpha).
\label{eq:model}
\end{aligned}
\end{equation}
Following exponential family notation, $\Ell(\cdot)$ is a likelihood function, $g(\cdot)$ is an invertible link function that maps the real numbers onto the likelihood's support, and $\btheta$ are other likelihood-specific parameters. 
Following~\cite{wilson2013gaussian} and \cite{oliva2016bayesian}, we assume $p(\w)$ is a Dirichlet process mixture of Gaussians~\citep[DP-GMM,][]{ferguson1973bayesian,antoniak1974mixtures}. 
By sampling from the posterior of $\w$, we can explore the space of stationary kernels and estimate the kernel hyperparameters in a Bayesian way. We assign each $\w_m$ in $\smash{\W = [\w_1 \dots \w_{M/2}]^{\top}}$ to a mixture component with the variable $z_m$, which is distributed according to a Chinese restaurant process~\citep[CRP,][]{aldous1985exchangeability} with concentration parameter $\alpha$. This prior introduces additional random variables: the mixture means $\smash{\{ \bmu_k \}_{k=1}^{K}}$, and the mixture covariance matrices $\smash{\{ \bSigma_k \}_{k=1}^{K}}$ where $K$ is the number of clusters in the current Gibbs sampling iteration.


The randomized map in Eq. \ref{eq:rff_def} allows us to approximate the original GPLVM in Eq. \ref{eq:gplvm_def} as
\begin{equation}
\y_j \sim \N_N(\bPhi \bbeta_j, \sigma^{2}_j \eye),
\quad
\bbeta_j \sim \N_M(\b_0, \B_0),
\quad
\x_n \sim \N_{D}(\zero, \eye).
\label{eq:gaussian_rflvm_def}
\end{equation}
We approximate $f_j(\X)$ in Eq. \ref{eq:gplvm_def} as $\bPhi \bbeta_j$, where $\smash{\bPhi = [\varphi(\x_1) \dots \varphi(\x_N)]^{\top}}$. This is a Gaussian RFLVM when $\Ell(\cdot)$ is a Gaussian distribution and $g(\cdot)$ is the identity function. Because the prior distribution on the mapping weights $\bbeta_j$ is Gaussian, the model is analogous to Bayesian linear regression given $\bPhi$; if we integrate out $\bbeta_j$, we recover a marginal likelihood that approximates the GPLVM's marginal likelihood. 

We use this representation to generalize the RFLVM to other observation types in the exponential family. For example, a Poisson RFLVM takes the following form:
\begin{equation}
    \y_j \sim \poisson(\exp(\bPhi \bbeta_j)),
    \quad
    \bbeta_j \sim \N_M(\b_0, \B_0), 
    \quad
    \x_n \sim \N_D(\zero, \eye).
    \label{eq:poisson_rflvm}
\end{equation}
For distributions including the Bernoulli, binomial, and negative binomial, the functional form of the data likelihood is
\begin{equation}
    \mathcal{L}(\bPhi, \bbeta_j, a(\y_j), b(\y_j), c(\y_j))
    =
    \prod_{n=1}^{N}
    c(y_{nj})
    \frac{(\exp(\varphi(\x_n) \bbeta_j))^{a(y_{nj})}}{(1 + \exp(\varphi(\x_n) \bbeta_j))^{b(y_ {nj})}},
    \label{eqn:logistic}
\end{equation}
for some functions of the data $a(\cdot)$, $b(\cdot)$, and $c(\cdot)$. The general form of this logistic RFLVM is then:
\begin{equation}
    \y_j \sim \mathcal{L}(\bPhi, \bbeta_j, a(\y_j), b(\y_j), c(\y_j)),
    \quad
    \bbeta_j \sim \N_M(\b_0, \B_0),
    \quad
    \x_n \sim \N_D(\zero, \eye).
    \label{eq:logistic_rflvm}
\end{equation}
For example, by setting $a(y_{nj}) = y_{nj}$, $b(y_{nj}) = y_{nj} + r_j$, and $c(y_{nj}) = {y_{nj} + r_j - 1 \choose y_{nj}}$, we get the negative binomial RFLVM with feature-specific dispersion parameter $r_j$.
%
\subsection{Inference for RFLVMs}\label{sec:base_gibbs}
We now present a general Gibbs sampling framework for all RFLVMs. 
First, we write the Gibbs sampling steps to estimate the posterior of the covariance kernel. Next, we describe estimating the latent variable $\X$ by taking the MAP estimate. Then, we sample the data likelihood-specific parameters $\btheta$ and linear coefficients $\bbeta_j$. 
Variables subscripted with zero, e.g., $\theta_0$, denote hyperparameters. While the number of mixture components may change across sampling iterations, let $K$ denote the number of components in the current Gibbs sampling step. We initialize all the parameters in our model by drawing from the prior, except for $\X$, which we initialize with PCA.

First, we sample $z_m$ following Algorithm 8 from \cite{neal2000markov}. Let $\smash{n_k = \sum_{\ell} \delta(z_{\ell} = k)}$, and let $n_k^{-m}$ denote the same sum with $z_m$ excluded. Then we sample the posterior of $z_m$ from the following discrete distribution for $k=1, 2, \ldots, K$:
\begin{equation}
p(z_m = k \mid \bmu, \bSigma, \W, \alpha) = \begin{cases}
\frac{n_k^{-m}}{M - 1 + \alpha} \N(\w_m \mid \bmu_k, \bSigma_k) & n_k^{-m} > 0
\\
\frac{\alpha}{M - 1 + \alpha} \int \N(\w_m \mid \bmu, \bSigma) \text{NIW}(\bmu, \bSigma) \text{d}\bmu \text{d}\bSigma & n_k^{-m} = 0.
\end{cases}
\label{eq:zm_sampling}
\end{equation}
Given assignments $\z = [z_1 \dots z_{M/2}]^{\top}$ and RFFs $\W$, the posterior of $\bSigma_k$ is inverse-Wishart distributed. Given $\bSigma_k$, the posterior of $\bmu_k$ is normally distributed~\citep{gelman2013bayesian}:
\begin{equation}
\begin{aligned}
\bSigma_k &\sim \mathcal{W}^{-1}(\bPsi_k, \nu_k),
\quad
\bmu_k \sim \N(\mathbf{m}_k, \frac{1}{\lambda_k} \bSigma_k).
\\
\bPsi_k &= \bPsi_0 + \sum_{m:z_m = k}^{M/2} (\w_m - \bar{\w}^{(k)})(\w_m - \bar{\w}^{(k)})^{\top} + \frac{\lambda_0 n_k}{\lambda_0 + n_k} (\w_m - \bmu_0)(\w_m - \bmu_0)^{\top}
\\
\bar{\w}^{(k)} &= \frac{1}{n_k} \sum_{m:z_m=k}^{M} \w_m,
\quad
\nu_k = \nu_0 + n_k,
\quad
\mathbf{m}_k = \frac{\lambda_0 \bmu_0 + n_k \bar{\w}^{k}}{\lambda_0 + n_k},
\quad
\lambda_k = \lambda_0 + n_k.
\end{aligned}
\label{eq:mu_sigma_sampling}
\end{equation}
We cannot sample from the full conditional distribution of $\W$, but prior work suggested a Metropolis–Hastings (MH) sampler using proposal distribution $q(\W)$ set to the prior $p(\W \mid \z, \bmu, \bSigma)= \N_{D}(\bmu_{z_m}, \bSigma_{z_m})$ (Eq.~\ref{eq:model}) and acceptance ratio $\rho_{\texttt{MH}}$~\citep{oliva2016bayesian}:
\begin{equation}
\w_m^{\star} \sim q(\W) \triangleq p(\W \mid \z, \bmu, \bSigma),
\quad
\rho_{\texttt{MH}} = \min \Bigg\{1, \frac{p(\Y \mid \X, \w_m^{\star}, \btheta)}{p(\Y \mid \X, \w_m, \btheta)}\Bigg\}.
\label{eq:w_sampling}
\end{equation}
Finally, we sample the DP-GMM concentration parameter $\alpha$~\citep{escobar1995bayesian}. 
We augment the model with variable $\eta$ to make sampling $\alpha$ conditionally conjugate:
\begin{equation}
\begin{aligned}
    \eta &\sim \betadist(\alpha + 1, M),
    \quad
    \frac{\pi_{\eta}}{1-\pi_{\eta}} = \frac{a_\alpha + K - 1 }{M(b_\alpha - \log(\eta))},
    \quad
    K = \left| \{ k: n_k > 0 \} \right|,
    \\
    \alpha &\sim \pi_{\eta} \gammadist(a_\alpha + K, b_\alpha - \log(h)) + (1 - \pi_{\eta}) \gammadist(a_\alpha + K - 1, b_\alpha - \log(\eta)).
\end{aligned}
\label{eq_alpha_sampling}
\end{equation}
For the Gaussian RFLVM (Eq.~\ref{eq:gaussian_rflvm_def}), let $\B_0 = \sigma^{-2} \S_0$. We integrate out $\smash{\bbeta_j}$ and $\smash{\sigma^{-2}}$ in closed form to obtain a marginal likelihood,
\begin{equation}
    p(\y_j \mid \X, \W) = \frac{1}{(2 \pi)^{N/2}} 
    \cdot
    \sqrt{\frac{ \left| \S_0 \right|}{\left|\S_N\right|}}
    \cdot
    \frac{b_0^{a_0}}{b_N^{a_N}}
    \cdot
    \frac{\Gamma(a_N)}{\Gamma(a_0)},
    \label{eq:gaussian_log_marginal_likelihood}
\end{equation}
where $\S_N = \bPhi^{\top} \bPhi + \S_0$, $\bbeta_N = \S_N^{-1} (\bbeta_0^{\top} \S_0 + \bPhi^{\top} \y_j)$, $a_N = a_0 + N / 2$, and $b_N = b_0 + (1/2)(\y_j^{\top} \y_j + \bbeta_0^{\top}\S_0 \bbeta_0 - \bbeta_N^{\top} \S_N \bbeta_N)$. See Appendix~\ref{app:linbayes} or \cite{minka2000bayesian} for details. However, inference can be slow because marginalizing out $\smash{\bbeta_j}$ introduces dependencies between the latent variables, and the complexity becomes $\mathcal{O}(NM^2)$. Alternatively, we can Gibbs sample $\bbeta_j$ and take the MAP estimate of $\X$ using the original log likelihood where the computational complexity is $\mathcal{O}(NM)$. 

In the Poisson RFLVM (Eq.~\ref{eq:poisson_rflvm}), we no longer have the option of marginalizing out $\smash{\bbeta_j}$. Instead, we take iterative MAP estimates of $\bbeta_j$ and $\X$. Given $\bPhi$, inference for $\bbeta_j$ is analogous to Bayesian inference for a Poisson generalized linear model (GLM). In Secs.~\ref{sec:experiments:simulated_data} and~\ref{sec:experiments:hippo}, we show that, by inducing closed-form gradients with respect to $\X$ through RFFs, this iterative MAP procedure produces results that are competitive with specialized GP-based latent variable models on count data.

For logistic RFLVMs (Eq.~\ref{eq:logistic_rflvm}), we use \polya-gamma augmentation~\citep{polson2013bayesian} to make inference tractable. A random variable $\omega$ is \polya-gamma distributed with parameters $b > 0$ and $c \in \Reals$, denoted $\omega \sim \text{PG}(b, c)$, if
\begin{equation}
    \omega \stackrel{d}{=} \frac{1}{2\pi^2} \sum_{k=1}^{\infty} \frac{g_k}{(k-1/2)^2 + c^2/(4\pi^2)},
\end{equation}
where $\stackrel{d}{=}$ denotes equality in distribution and $g_k \sim \gammadist(b, 1)$ are independent gamma random variables. The identity critical for \polya-gamma augmentation is
\begin{equation}
\frac{(e^{\psi_{nj}})^{a_{nj}}}{(1 + e^{\psi_{nj}})^{b_{nj}}} = 2^{-b_{nj}} e^{\kappa_{nj} \psi_{nj}} \int_{0}^{\infty} e^{- \omega \psi_{nj}^2 / 2} p(\omega) \text{d}\omega,
\label{eq:pg_aug_integral}
\end{equation}
where $\kappa_{nj} = a_{nj} - b_{nj}/2$ and $p(\omega) = \text{PG}(\omega \mid b_{nj}, 0)$.  If we define $\psi_{nj} = \varphi(\x_n)^{\top} \bbeta_j$, then Eq.~\ref{eq:pg_aug_integral} allows us to rewrite the likelihood in Eq.~\ref{eqn:logistic} as proportional to a Gaussian. Furthermore, we can sample $\omega$ conditioned on $\psi_{nj}$ as $p(\omega \mid \psi_{nj}) \sim \text{PG}(b_{nj}, \psi_{nj})$. This enables convenient, closed-form Gibbs sampling steps of $\bbeta_j$, conditioned on \polya-gamma augmentation variables $\omega_{nj}$:
\begin{equation}
\begin{aligned}
    \omega_{nj} \mid \bbeta_j &\sim \text{PG}( b_{nj}, \varphi(\x_n)^{\top} \bbeta_j),
    &\quad \textbf{V}_{\bomega_j} &= (\bPhi^{\top} \bOmega_j \bPhi + \B_0^{-1})^{-1},
    \\
    \bbeta_j \mid \bOmega_j &\sim \N(\textbf{m}_{\bomega_j}, \textbf{V}_{\bomega_j}),
    &\quad \textbf{m}_{\bomega_j} &= \textbf{V}_{\bomega_j} (\bPhi^{\top} \bkappa_j + \B_0^{-1} \bbeta_0),
\end{aligned}
\end{equation}
where $\bOmega_j = \text{diag}([\omega_{1j} \dots \omega_{Nj}])$ and $\bkappa_j = [\kappa_{1j} \dots \kappa_{Nj}]^{\top}$. This technique has been used to derive Gibbs samplers for binomial regression \citep{polson2013bayesian}, negative binomial regression \citep{zhou2012lognormal}, and correlated topic models \citep{chen2013scalable, linderman2015dependent}. Here, we use it to derive a sampling approach for logistic RFLVMs.

RFLVMs are identifiable up to the rotation and scale of the latent variable $\X$. As a result, MAP estimates of $\X$ between iterations are unaligned as they can be arbitrarily rescaled and rotated through inference. Thus, a point estimate of $\X$ that is a function of the Monte Carlo samples of $\X$, e.g. the expectation of $\X$ across the samples, will not be meaningful. To this end, we arbitrarily fix the rotation of $\X$ by taking the singular value decomposition (SVD) of the MAP estimate, $\hat{\X} = \mathbf{USV}^{T}$, and setting $\X$ to be the left singular vectors corresponding to the $D$ largest singular values, $\X  \triangleq \left[ \mathbf{u}_{1}, \ldots , \mathbf{u}_{D} \right] $ where $\mbox{diag}(\mathbf{S}) = [s_1 \,\ldots\, s_D]$ and $s_1 \geq s_2 \geq \ldots \geq s_D$. Then we rescale $\X$ so that the covariance of the latent space is the identity matrix. This has the effect of enforcing orthogonality, and does not allow heteroskedasticity in the latent dimensions.

%% file: 3_experiments.tex
\section{Experiments}\label{sec:experiments}
\begin{figure}[t!]
\centering
\includegraphics[width=\linewidth]{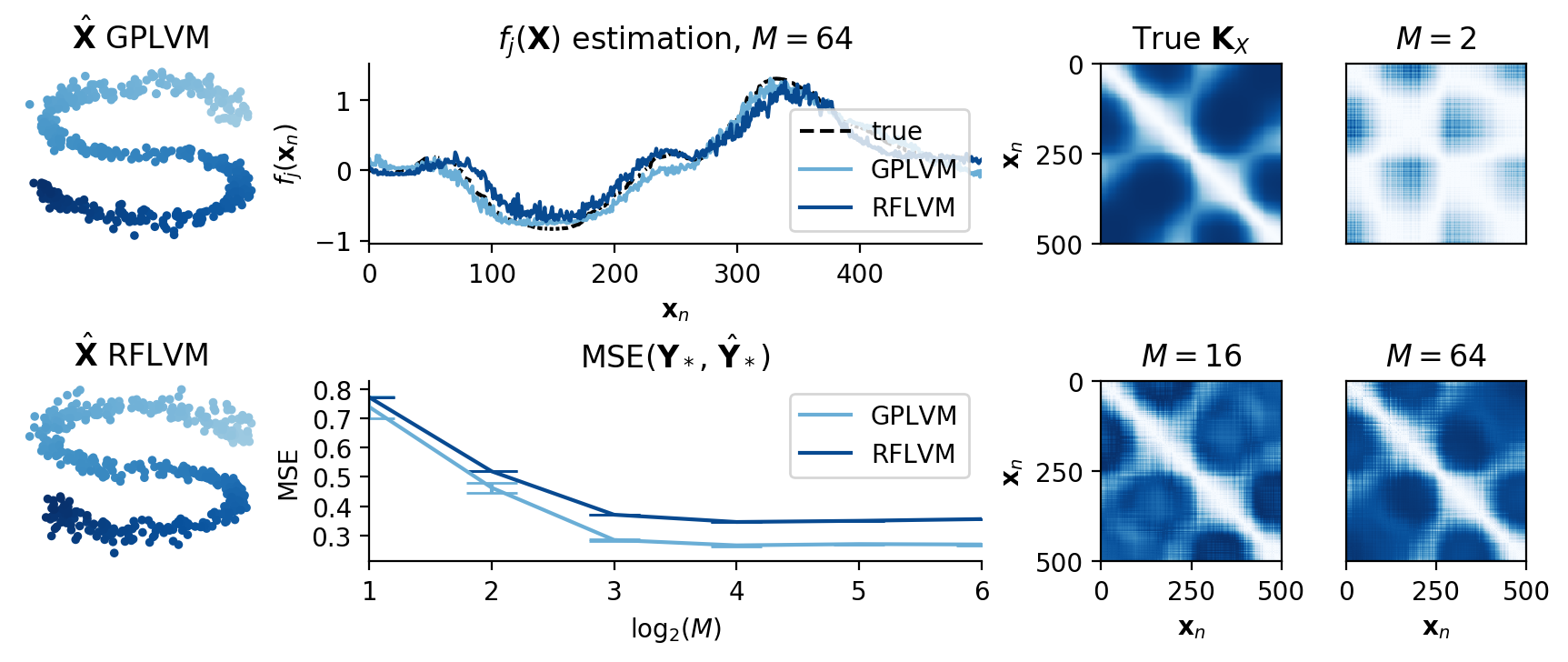}
\caption{
    \textbf{Simulated data with Gaussian emissions.} (Left) Inferred latent variables for both a GPLVM and Gaussian RFLVM. (Upper middle) Comparison of estimated $f_j(\X)$ for a single feature as estimated by GPLVM and RFLVM. (Lower middle) Comparison of MSE reconstruction error on held out $\Y_{*}$ for increasing $M$, where $M$ is the number of inducing points for GPLVM and random Fourier features for RFLVM. (Right) Ground truth covariance matrix $\K_X$ compared with RFLVM estimation for an increasing number of random Fourier features $M$.
}
\label{fig:gaussian}
\end{figure}
In our results, we refer to the Gaussian-distributed GPLVM using inducing point methods for inference as \emph{GPLVM}~\citep{titsias2010bayesian}. %
We fit all GPLVM experiments using the \texttt{GPy} package~\citep{gpy2014}. We refer to the Poisson-distributed GPLVM using a double Laplace approximation as \emph{DLA-GPLVM}~\citep{wu2017gaussian}. DLA-GPLVM is designed to model multi-neuron spike train data, and the code\footnote{\url{https://github.com/waq1129/LMT}} initializes the latent space using the output of a Poisson linear dynamical system \citep{macke2011empirical}, and places a GP prior on $\X$. To make the experiments comparable for all GPLVM experiments, we initialize DLA-GPLVM with PCA and assume $\x_n \sim \N_{D}(\zero, \eye)$. 
We refer to our GPLVM with random Fourier features as \emph{RFLVM} and explicitly state the assumed distribution. In Sec. \ref{sec:experiments:simulated_data}, we use a Gaussian RFLVM with the linear coefficients $\{ \bbeta_j \}_{j=1}^{J}$ marginalized out (Eq. \ref{eq:gaussian_log_marginal_likelihood}) for a fairer comparison with the GPLVM. In Sec. \ref{sec:experiments:image}, we use a Gaussian RFLVM without marginalizing out the linear coefficients because inference was faster on the larger datasets.

Since hyperparameter tuning our model on each dataset would be both time-consuming and unfair without also tuning the baselines, we fixed the hyperparameters across experiments. We used 2000 Gibbs sampling iterations with $1000$ burn-in steps, $M = 100$, and $D = 2$. We initialized $K = 20$ and $\alpha = 1$. In Sec.~\ref{sec:experiments:hippo}, we used $D=3$ and visualized $\hat{\X}$ after the best affine transformation onto the 2-D rat positions following \cite{wu2017gaussian}. For computational reasons, MNIST and CIFAR-10 were subsampled (see Appendix~\ref{app:experiments} for details).

\subsection{Simulated data}\label{sec:experiments:simulated_data}
We first evaluate RFLVM on simulated data. 
We set $\X$ to be a two-dimensional S-shaped manifold, sampled functions $\F = \smash{\{ f_j(\X) \}_{j=1}^{J}}$ from a Gaussian process with an RBF kernel, and then generated observations for Gaussian emissions (Eq. \ref{eq:gplvm_def}) and Poisson emissions (Eq. \ref{eq:poisson_rflvm}). For all simulations, we used $N = 500, J = 100$, and $D = 2$. 

For these experiments, we computed the mean-squared error (MSE) between test set observations, $\Y_{*}$, and predicted observations $\smash{\hat{\Y}_{*}}$, where we held out 20\% of the observations for the test set. To evaluate our latent space results, we projected the estimated latent space, $\smash{\hat{\X}}$, onto the hyperplane that minimized the squared error with the ground truth, $\smash{\lVert \X - \hat{\X}\mathbf{A} \rVert_{2}^{2}}$, and calculated the $R^2$ value between the true $\X$ and the projected latent space $\smash{\hat{\X} \mathbf{A}}$. We evaluated our model's ability to estimate the GP outputs $\smash{f_j(\X) \approx \varphi(\X) \bbeta_j}$ by comparing the MSE between the estimated $\smash{\varphi(\hat{\X}) \bbeta_j}$ and the true generating $f_j(\X)$. We computed the mean and standard error of the MSE and $R^2$ results by running each experiment five times. 

We compared the performance of a Gaussian RFLVM to the GPLVM. We ran these experiments across multiple values of $M$, where $M$ denotes the number of random features for the RFLVM and the number of inducing points for the GPLVM. Both models recovered the true latent variable $\X$ accurately and estimated the nonlinear maps, $\F$, well (Fig.~\ref{fig:gaussian}, upper middle). Empirically, a GPLVM shows better performance for estimating $\Y_{*}$ than the RFLVM (Fig.~\ref{fig:gaussian}, lower middle). We hypothesize that this is because \nystrom's method has better generalization error bounds than RFFs when there is a large gap in the eigenspectrum~\citep{yang2012nystrom}, which is the case for $\K_X$.
However, we see that the RFLVM approximates the true $\K_X$ given enough random features (Fig.~\ref{fig:gaussian}, right), though perhaps less accurately than the GPLVM (Fig.~\ref{fig:gaussian}, lower middle).

\begin{figure}[t!]
\centering
\includegraphics[width=\linewidth]{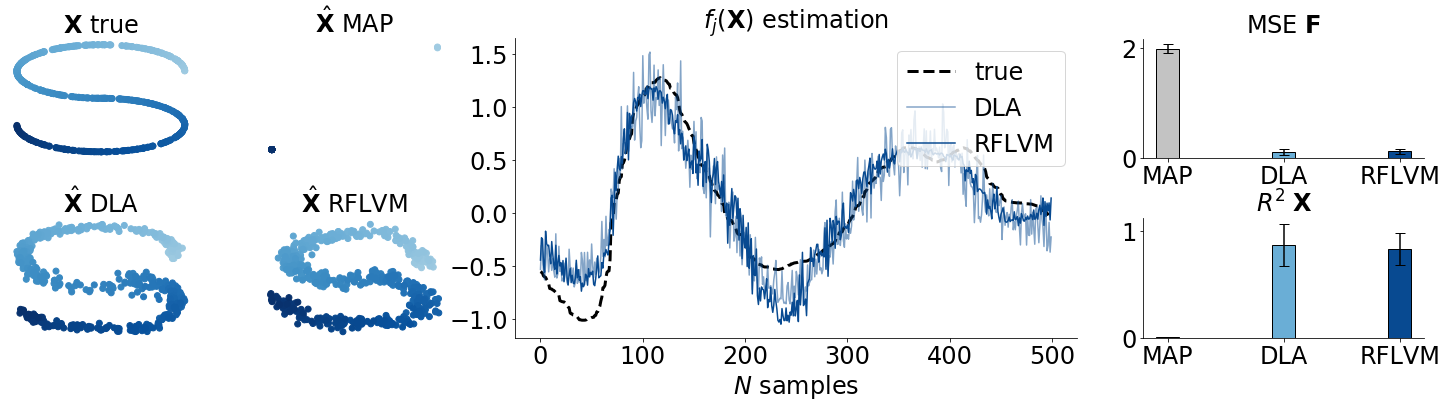}
\caption{
    \textbf{Simulated data with Poisson emissions.}
    (Left four plots) The true latent variable $\X$ compared with \smash{$\hat{\X}$} estimated using a MAP-GPLVM, a DLA-GPLVM, and a Poisson RFLVM. (Middle) Comparison of \smash{$f_j(\X)$} for a single feature as estimated by DLA-GPLVM and RFLVMs. (Right) MSE and $R^2$ between the true $\F$ and \smash{$\hat{\F}$}, and the true $\X$ and \smash{$\hat{\X}$}, respectively.
}
\label{fig:poisson}
\label{fig:illustration}
\end{figure}

To demonstrate the utility of our model beyond Gaussian distributed data, we compared results for count data from a Poisson RFLVM and a DLA-GPLVM. Additionally, we compared results to our own naive implementation of the Poisson GPLVM that performs coordinate ascent on $\X$ and $\F$ by iteratively taking MAP estimates without using RFFs. We refer to this method as \emph{MAP-GPLVM}. 
The MAP-GPLVM appears to get stuck in poor local modes~\citep{wu2017gaussian} because we do not have gradients of the posterior in closed form (Fig.~\ref{fig:poisson}, left). Both DLA-GPLVM and RFLVM, however, do have closed-form gradients and approximate the true manifold with similar $R^2$ and MSE values for $\smash{\hat{\X}}$ and $\smash{\hat{f}_j(\X)}$. 

\subsection{Hippocampal place cell data}\label{sec:experiments:hippo}
\begin{figure}[t!]
\centering
\includegraphics[width=\linewidth]{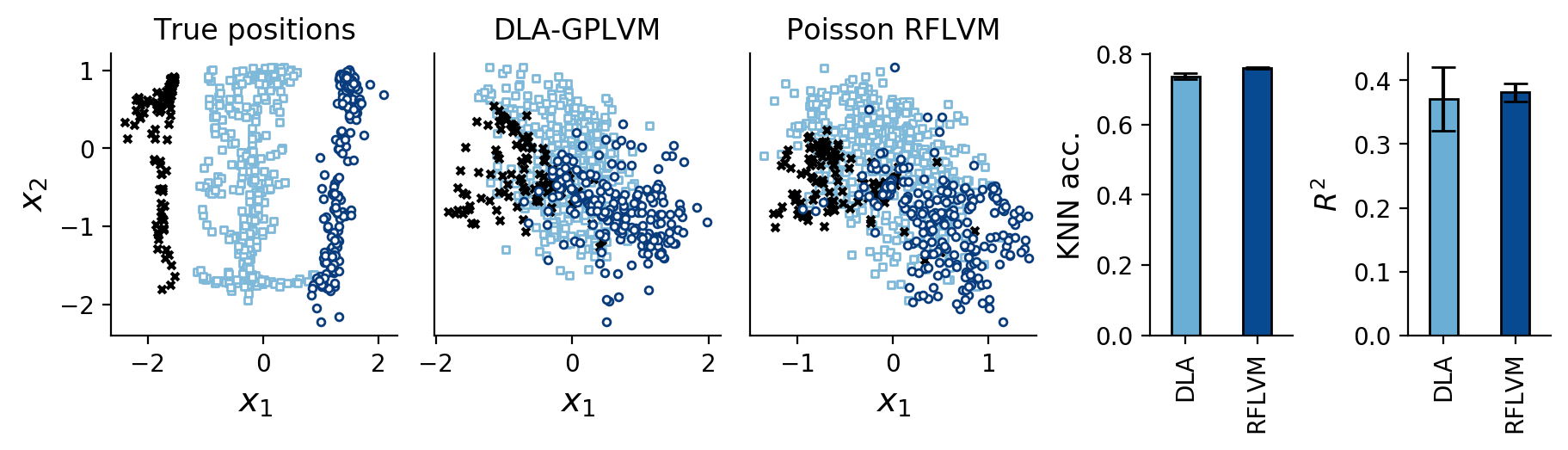}
\caption{
    \textbf{Hippocampal place cells.}
    (Left three plots) Inferred latent space for the DLA-GPLVM and the Poisson RFLVM. The points are colored by three major regions of the true rat position in a W-shaped maze. (Right two plots) KNN accuracy using 5-fold cross validation and $\smash{R^2}$ performance of the best affine transformation from $\smash{\hat{\X}}$ onto the rat positions $\X$. Error bars computed using five trials.
}
\label{fig:hippo}
\end{figure}

\begin{figure}[t!]
\centering
\includegraphics[width=\linewidth]{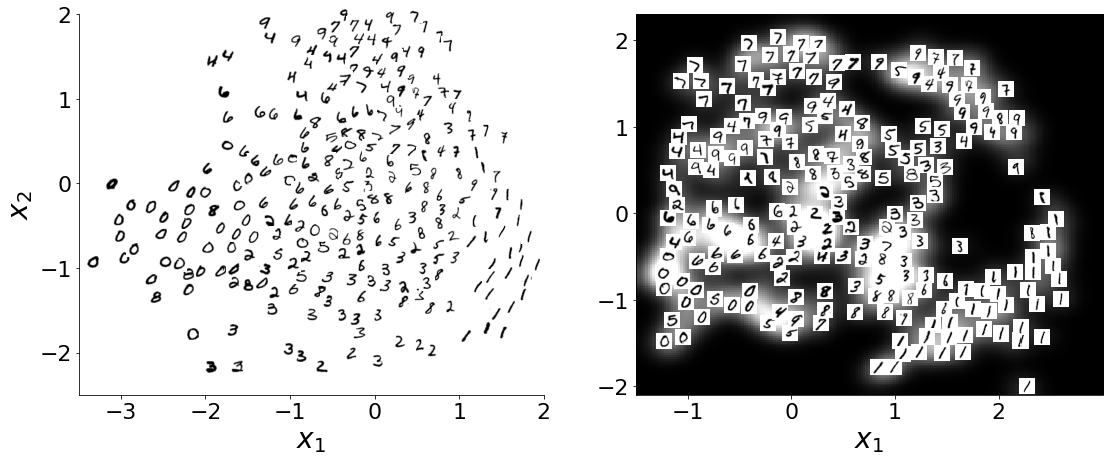}
\caption{
    \textbf{MNIST digits.}
    Digits visualized in 2-D latent space inferred from DLA-GPLVM (left) and Poisson RFLVM (right). Following \cite{lawrence2004gaussian}, we plotted images in a random order while not plotting any images that result in an overlap. The RFLVM's latent space is visualized as a histogram of 1000 draws after burn-in. The plotted points are the sample posterior mean.
}
\label{fig:mnist}
\end{figure}
Next, we checked whether a non-Gaussian RFLVM recovers an interpretable latent space when applied to a scientific problem. 
In particular, we use an RFLVM to model hippocampal place cell data \citep{wu2017gaussian}. Place cells, a type of neuron, are activated when an animal enters a particular place in its environment. Here, $\Y$ is an $N \times J$ matrix of count-valued spikes where $n$ indexes time and $j$ indexes neurons. These data were jointly recorded while measuring the position of a rat in a W-shaped maze. We are interested in reconstructing the latent positions of the rat with $\X$. 

We quantified goodness-of-fit of the latent space by assessing how well the RFLVM captures known structure, in the form of held-out sample labels, in the low-dimensional space. 
After estimating $\smash{\hat{\X}}$, we performed $K$-nearest neighbors (KNN) classification on $\smash{\hat{\X}}$ with $K=1$. We ran this classification five times using 5-fold cross validation. We report the mean and standard deviation of KNN accuracy across five of these experiments. 

The Poisson RFLVM and DLA-GPLVM have similar performance in terms of how well they cluster samples in the latent space as measured by KNN accuracy using regions of the maze as labels. Furthermore, the models have similar performance in recovering the true rat positions $\X$, measured by $R^2$ performance (Fig.~\ref{fig:hippo}). These results suggest that our generalized RFLVM framework finds structure even in empirical, complex, non-Gaussian data and is competitive with models built for this specific task. 

\subsection{Text and image data}\label{sec:experiments:image}

Finally, we examine whether an RFLVM captures the latent space of text, image, and empirical data sets. We hold out the labels and use them to evaluate the estimated latent space using the same KNN evaluation from Sec.~\ref{sec:experiments:hippo}.
Across all eight data sets, the Poisson and negative binomial RFLVMs infer a low-dimensional latent variable $\smash{\hat{\X}}$ that generally captures the latent structure as well as or better than linear methods like PCA and NMF \citep{lee1999learning}. Moreover, adding nonlinearity but retaining a Gaussian data likelihood---as with real-valued models like Isomap \citep{tenenbaum2000global}, a variational autoencoder \citep[VAE,][]{kingma2013auto}, and the Gaussian RFLVM, or even using the Poisson-likelihood DLA-GPLVM---perform worse than the Poisson and negative binomial RFLVMs (Tab.~\ref{tab:real_dataset_results}, Fig.~\ref{fig:mnist}). We posit that this improved performance is because the generating process from the latent space to the observations for these data sets is (in part) nonlinear, non-RBF, and integer-valued. 
\begin{table}[h]
\caption{Classification accuracy evaluated by fitting a KNN classifier ($K = 1$) with five-fold cross validation.  Mean accuracy and standard error were computed by running each experiment five times.}
\label{table:clustering_results}
\begin{center}
\resizebox{\columnwidth}{!}{%
\begin{tabular}{lcccc}
\toprule
& PCA & NMF & Isomap & VAE \\
\midrule
Bridges     & $0.8469 \pm 0.0067$ & $\mathbf{0.8664 \pm 0.0164}$ & $0.8375 \pm 0.0240$ & $0.8141 \pm 0.0301$ \\
CIFAR-10    & $0.2651 \pm 0.0019$ & $0.2450 \pm 0.0028$ & $0.2716 \pm 0.0056$ & $0.2711 \pm 0.0083$ \\
Congress    & $0.5558 \pm 0.0098$ & $0.5263 \pm 0.0108$ & $0.5239 \pm 0.0178$ & $0.6563 \pm 0.0314$ \\
MNIST       & $0.3794 \pm 0.0146$ & $0.2764 \pm 0.0197$ & $0.4408 \pm 0.0192$ & $\mathbf{0.6512 \pm 0.0228}$ \\
Montreal    & $0.6802 \pm 0.0099$ & $0.6878 \pm 0.0207$ & $0.7049 \pm 0.0098$ & $0.6702 \pm 0.0325$ \\
Newsgroups  & $0.3896 \pm 0.0043$ & $0.3892 \pm 0.0042$ & $0.4021 \pm 0.0098$ & $0.3926 \pm 0.0113$ \\
Spam        & $0.8454 \pm 0.0037$ & $0.8237 \pm 0.0040$ & $0.8272 \pm 0.0047$ & $0.9028 \pm 0.0128$ \\
Yale        & $0.5442 \pm 0.0129$ & $0.4739 \pm 0.0135$ & $0.5891 \pm 0.0155$ & $0.6327 \pm 0.0209$ \\
\midrule
& DLA-GPLVM & Gaussian RFLVM & Poisson RFLVM & Neg. binom. RFLVM \\
\midrule
Bridges     & $0.8578 \pm 0.0101$ & $0.8512 \pm 0.0134$ & $0.8440 \pm 0.0165$ & $\mathbf{0.8664 \pm 0.0191}$ \\
CIFAR-10    & $0.2641 \pm 0.0063$ & $0.2755 \pm 0.0132$ & $\mathbf{0.2789 \pm 0.0080}$ & $0.2656 \pm 0.0048$ \\
Congress    & $0.7815 \pm 0.0185$ & $0.5693 \pm 0.0107$ & $0.7673 \pm 0.0109$ & $\mathbf{0.8093 \pm 0.0154}$ \\
MNIST       & $0.3820 \pm 0.0121$ & $0.5569 \pm 0.0503$ & $0.6494 \pm 0.0210$ & $0.4463 \pm 0.0313$ \\
Montreal    & $0.2885 \pm 0.0001$ & $0.7533 \pm 0.0165$ & $\mathbf{0.8158 \pm 0.0210}$ & $0.7530 \pm 0.0478$ \\
Newsgroups  & $0.3687 \pm 0.0077$ & $0.4006 \pm 0.0083$ & $\mathbf{0.4144 \pm 0.0029}$ & $0.4045 \pm 0.0044$ \\
Spam        & $\mathbf{0.9521 \pm 0.0069}$ & $0.8616 \pm 0.0051$ & $0.9515 \pm 0.0023$ & $0.9443 \pm 0.0035$ \\
Yale        & $0.4788 \pm 0.0991$ & $0.6179 \pm 0.0092$ & $\mathbf{0.6894 \pm 0.0295}$ & $0.5394 \pm 0.0117$ \\
\bottomrule
\end{tabular}
}
\end{center}
\label{tab:real_dataset_results}
\end{table}

%% file: 4_conclusion.tex
\section{Conclusion}\label{sec:conclusion}
We presented a framework that uses random Fourier features to induce computational tractability between the latent variables and GP-distributed maps in Gaussian process latent variable models. Our approach allows the Gaussian model to be extended to arbitrary distributions, and we derived an RFLVM for Gaussian, Poisson and logistic distributions. We described distribution-specific inference techniques for each posterior sampling step. 
Our empirical results showed that each was competitive in downstream analyses with existing distribution-specific approaches on diverse data sets including synthetic, image, text, and multi-neuron spike train data. We are particularly interested in exploring extensions of our generalized RFLVM framework to more sophisticated models such as extending GP dynamic state-space models \citep{ko2011learning} to count data and neuroscience applications, which assume temporal structure in $\X$.

RFLVMs have a number of limitations that motivate future work. First, the latent variables are unidentifiable up to scale and rotation. Our rescaling procedure (Sec.~\ref{sec:base_gibbs}) does not allow heteroscedastic dimensions and enforces orthogonality between the Gaussian latent variables. This prevents the use of more structured priors, such as a GP prior on $\X$, since any inferred structure is eliminated between iterations. We are interested in adopting constraints from factor analysis literature to address the identifiability issues without a restrictive rescaling procedure~\citep{erosheva2011dealing, millsap2001trivial,ghosh2009default}. Second, label switching in mixture models is a well-studied challenge that is present in our model. Enforcing identifiability may improve inference and model interpretability~\citep{stephens2000dealing}. Finally, our model has a number of hyperparameters such as the latent dimension, the number of random Fourier features, and the number of Gibbs sampling iterations. Both simplifying the model and estimating these hyperparameters from data are two important directions to improve the usability of RFLVMs.



%% file: 0_acknowledgements.tex

%% file: 6_appendix.tex
\appendix
 
\section{Experiments}\label{app:experiments}

\subsection{Additional results}

\begin{figure}[h]
\centering
\includegraphics[width=\linewidth]{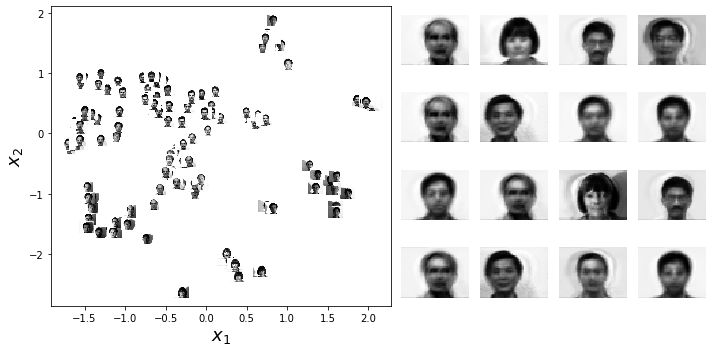}
\caption{Latent space and generated faces for the Yale dataset using a Poisson RFLVM.}
\end{figure}

\begin{figure}[h]
\centering
\includegraphics[width=\linewidth]{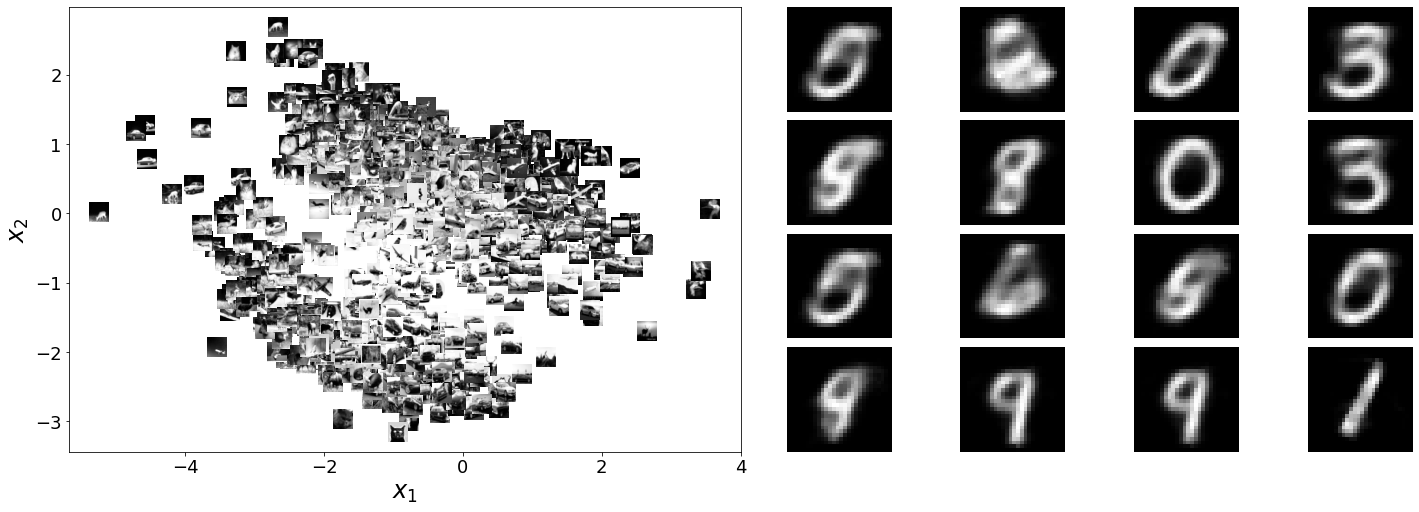}
\caption{Latent space for CIFAR-10 and generated digits for MNIST using a Poisson RFLVM.}
\end{figure}

\subsection{Data descriptions and preprocessing}\label{app:ds_preprocessing}

\begin{figure}[h]
\centering
\includegraphics[width=\linewidth]{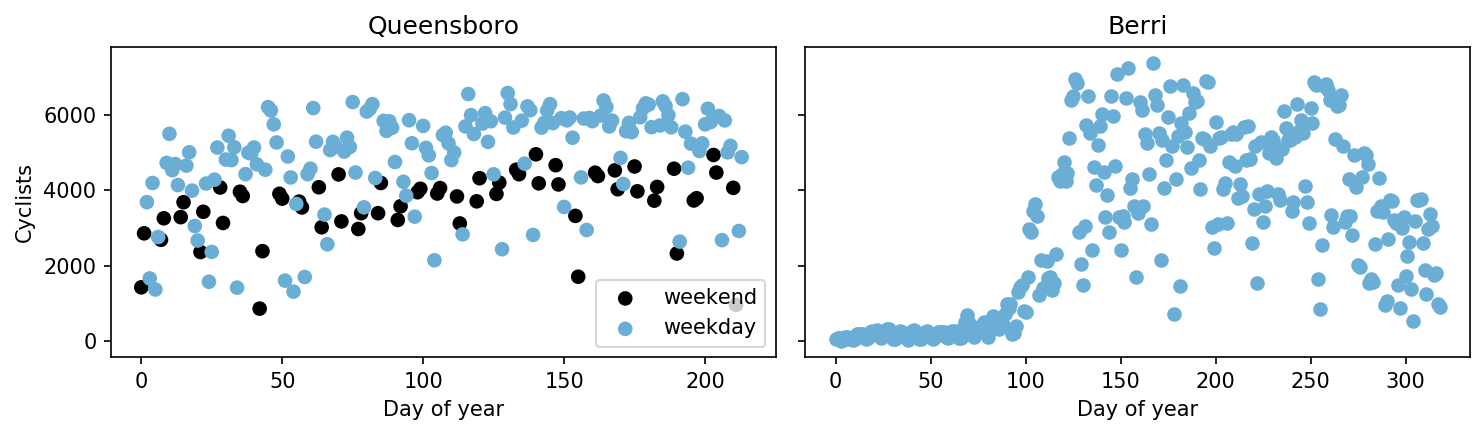}
\caption{(Left) The number of bicycle crossings over the Queensboro Bridge from April through November 2017. (Right) The number of cyclists on Berri St. in Montreal throughout 2015.}
\label{fig:bridges_montreal}
\end{figure}

\begin{itemize}
    \item{
        \textbf{Bridges:} We used the number of bicycle crossing per day over four East River bridges in New York City\footnote{\url{https://data.cityofnewyork.us/Transportation/Bicycle-Counts-for-East-River-Bridges/gua4-p9wg}}. Since these data are unlabeled, we used weekday vs. weekend as binary labels since such information is correlated with bicycle counts (Fig. \ref{fig:bridges_montreal}, left).
    }
    
    \item{
        \textbf{CIFAR-10:} We limited the classes to $[1-5]$ and subampled $400$ images for each class for a final dataset of size $2000$. We converted the images to grayscale and resized them from $32 \times 32$ down to $20 \times 20$ pixels.
    }
    
    \item{
        \textbf{Congress:} The word frequency counts from individual members of the 109th Congress~\citep{gentzkow2010drives}. Labels are political party: \textit{Democrat}, \textit{Independent}, \textit{Republican}.
        }
    
    \item{
        \textbf{MNIST:} We limited the dataset size by randomly subsampling 1000 images.
    }
    
    \item{
        \textbf{Montreal:} We use the number of cyclists per day on eight bicycle lanes in Montreal.\footnote{\url{http://donnees.ville.montreal.qc.ca/dataset/f170fecc-18db-44bc-b4fe-5b0b6d2c7297/resource/64c26fd3-0bdf-45f8-92c6-715a9c852a7b}}. Since these data are unlabeled, we used the four seasons as labels, since seasonality is correlated with bicycle counts (Fig. \ref{fig:bridges_montreal}, right).
    }
    
    \item{
        \textbf{Newsgroups:} The 20 Newsgroups Dataset\footnote{\url{http://qwone.com/~jason/20Newsgroups/}}. We limited the classes to \textit{comp.sys.mac.hardware}, \textit{sci.med}, and \textit{alt.atheism}, and limited the vocabulary to words with document frequencies in the range $10-90\%$.
    }
    
    \item{
        \textbf{Spam:} The SMS Spam dataset from the UCI Machine Learning Repository\footnote{\url{https://archive.ics.uci.edu/ml/datasets/SMS+Spam+Collection}}. Emails are labeled \textit{spam} or \textit{ham} (not spam).
    }
    
    \item{
        \textbf{Yale:} The Yale Faces Dataset\footnote{\url{http://vision.ucsd.edu/content/yale-face-database}}. We used subject IDs as labels.
    }
\end{itemize}

\subsection{Experiment details}

\textbf{GPLVM baselines:} We used GPy's implementation \texttt{BayesianGPLVMMiniBatch}, which supports inducing points and prediction on held out data.

\section{Marginal likelihood in Bayesian linear regression}\label{app:linbayes}

For ease of notation, we drop the $j$ subscript, and therefore $\Y \rightarrow \y = [y_1 \dots y_N]^{\top}$ and $\bbeta_j \rightarrow \bbeta$. Consider the linear regression model
\begin{equation}
\y = \bPhi \bbeta^{\top} + \bvarepsilon, \qquad \bvarepsilon \sim \N_N(\zero, \sigma^{2} \eye),
\end{equation}
where $\bPhi = [\varphi(\x_1) \dots \varphi(\x_N)]^{\top}$, an $N \times M$ matrix. A common conjugate prior on $\bbeta$ is a normal–inverse–gamma distribution,
\begin{equation}
\begin{aligned}
\bbeta \mid \sigma^2 &\sim \N_M(\bbeta_0, \sigma^2 \S_0^{-1})
\\
\sigma^2 &\sim \text{InvGamma}(a_0, b_0),
\end{aligned} \label{eq:linbayes_prior}
\end{equation}
We can write the functional form of the posterior and prior terms in \eqref{eq:linbayes_prior} as
\begin{equation}
\begin{aligned}
p(\y \mid \bPhi, \bbeta, \sigma^2)
&= (2 \pi \sigma^2)^{-N/2} \exp\!\Big(\!-\frac{1}{2 \sigma^2} (\y - \bPhi \bbeta)^{\top} (\y - \bPhi \bbeta) \Big)
\\
p(\bbeta \mid \sigma^2) &= (2 \pi \sigma^2)^{-M/2} \big|\S_0 \big|^{1/2} \exp\Big(\!-\frac{1}{2 \sigma^2} (\bbeta - \bbeta_0)^{\top} \S_0 (\bbeta - \bbeta_0) \Big)
\\
p(\sigma_{\beta}^2) &= \frac{b_0^{a_0}}{\Gamma(a_0)} (\sigma_{\beta}^2)^{-(a_0 + 1)} \exp\Big(\frac{-b_0}{\sigma^2}\Big).
\end{aligned}
\end{equation}
We can combine the likelihood's Gaussian kernel with the prior's kernel in the following way:
\begin{equation}
\begin{aligned}
&(\y - \bPhi \bbeta)^{\top} (\y - \bPhi \bbeta) + (\bbeta - \bbeta_0)^{\top} \S_0 (\bbeta - \bbeta_0)
\\
&= \y^{\top} \y
+
\bbeta_0^{\top} \S_0 \bbeta_0
-
\bbeta_N^{\top} \S_N \bbeta_N
+
(\bbeta - \bbeta_N)^{\top} \S_N (\bbeta - \bbeta_N).
\end{aligned}
\end{equation}
where $\bbeta_N$ and $\S_N$ are defined as
\begin{equation}
\begin{aligned}
\S_N &= \bPhi^{\top} \bPhi + \S_0
\\
\bbeta_N &= \S_N^{-1} (\bbeta_0^{\top} \S_0 + \bPhi^{\top} \y).
\end{aligned}
\label{eq:bn_bn_defs}
\end{equation}
Now our posterior can be written as
\begin{equation}
\begin{aligned}
p(\y \mid \bPhi, \bbeta, \sigma^2)
&\propto (2 \pi)^{-M/2} \big|\S_0 \big|^{1/2} \exp\Big(-\frac{1}{2\sigma^2}\Big[(\bbeta - \bbeta_N)^{\top} \S_N (\bbeta - \bbeta_N)\Big]\Big)
\\
&\;\;\;\; (2 \pi \sigma^2)^{-N/2} \exp\Big(-\frac{1}{2 \sigma^2}\Big[
\y^{\top} \y
+
\bbeta_0^{\top} \S_0 \bbeta_0
-
\bbeta_N^{\top} \S_N \bbeta_N
\Big]\Big)
\\
&\;\;\;\; \frac{b_0^{a_0}}{\Gamma(a_0)} (\sigma^2)^{-(a_0 + 1)} \exp\Big(\frac{-b_0}{\sigma^2}\Big).
\end{aligned}
\end{equation}
We can see that we have an $M$-variate normal distribution on the first line. If we ignore $(2 \pi)^{-N/2}$ and inverse–gamma prior normalizer, we can combine the bottom two lines to be proportional to an inverse–gamma distribution,
\begin{equation}
(\sigma^2)^{-(a_0 + N/2 + 1)}
\exp\Big(-\frac{1}{\sigma^2}\Big[
b_0
+
\frac{1}{2} \Big\{\y^{\top} \y
+
\bbeta_0^{\top} \S_0 \bbeta_0
-
\bbeta_N^{\top} \S_N \bbeta_N
\Big\}\Big]\Big).
\end{equation}

Now define $a_N$ and $b_N$ as
\begin{equation}
\begin{aligned}
a_N &= a_0 + \frac{N}{2}
\\
b_N &= b_0 + \frac{1}{2}(\y^{\top} \y + \bbeta_0^{\top}  \S_0 \bbeta_0 - \bbeta_N^{\top} \S_N \bbeta_N).
\end{aligned}
\label{eq:an_bn_defs}
\end{equation}
Thus, we can write our posterior as
\begin{equation}
\begin{aligned}
p(\bbeta, \sigma^2 \mid \bPhi, \y)
&\propto
p(\bbeta \mid \bPhi, \y_j, \sigma^2)
p(\sigma^2 \mid \bPhi, \y)
\\
&\!\!\!\!\text{where}
\\
\bbeta \mid \bPhi, \y, \sigma^2 &\sim \N_M(\bbeta_N, \S_N)
\\
\sigma^2 \mid \y, \bPhi &\sim \text{InvGamma}(a_N, b_N).
\end{aligned}
\end{equation}
Now to compute the log marginal likelihood, we want
\begin{equation}
    p(\y \mid \bPhi, a_0, b_0) = \int\!\!\!\int p(\y \mid \X, \bbeta, \sigma^2) p(\bbeta), \sigma^2 \mid a_0, b_0) \,\text{d}^M \bbeta \,\text{d}\sigma^2.
    \label{eq:marginal_likelihood}
\end{equation}
Using the definitions in \eqref{eq:bn_bn_defs} and \eqref{eq:an_bn_defs}, we can write the joint as
\begin{equation}
\begin{aligned}
p(\y, \bbeta, \sigma^2)
&=
(2 \pi \sigma^2)^{-P/2} \big|\S_0 \big|^{1/2} \exp\!\Big(\!-\frac{1}{2\sigma^2}\Big[(\bbeta - \bbeta_N)^{\top} \S_N (\bbeta - \bbeta_N)\Big]\Big)
\\
&\;\;\;\; (\sigma^2)^{-(a_N + 1)}
\exp\!\Big(\!-\frac{b_N}{\sigma^2}\Big)
\\
&\;\;\;\; (2 \pi)^{-N/2} \frac{b_0^{a_0}}{\Gamma(a_0)}.
\end{aligned}
\label{eq:linbayes_posterior_joint}
\end{equation}
The integral over $\bbeta$ is only over the Gaussian kernel, which allows us to compute it immediately:
\begin{equation}
(2\pi\sigma^2)^{M/2} \big|\S_N\big|^{-1/2}
=
\int \exp\Big(
-\frac{1}{2} (\bbeta - \bbeta_N)^{\top} \Big[ \frac{1}{\sigma^2} \S_N \Big] (\bbeta - \bbeta_N) \Big) \text{d}^M \bbeta.
\end{equation}
The terms $(2 \pi \sigma^2)^{M/2}$ in \eqref{eq:linbayes_posterior_joint} cancel, and the first line of \eqref{eq:linbayes_posterior_joint} reduces to
\begin{equation}
    \sqrt{\frac{|\S_0|}{|\S_N|}}.
\end{equation}
We can compute the second integral in \eqref{eq:marginal_likelihood} because we know the normalizing constant of the gamma kernel,
\begin{equation}
    \frac{\Gamma(a_N)}{b_N^{a_N}}
    =
    \int (\sigma^2)^{-(a_N + 1)} \exp\!\Big(\!-\frac{b_N}{\sigma^2}\Big) \text{d}\sigma^2.
\end{equation}
Putting everything together, we see that the marginal likelihood is
\begin{equation}
    p(\y \mid \bPhi, a_0, b_0) = \frac{1}{(2 \pi)^{N/2}} \cdot \sqrt{\frac{|\S_0 |}{|\S_N |}} \cdot \frac{b_0^{a_0}}{b_N^{a_N}} \cdot \frac{\Gamma(a_N)}{\Gamma(a_0)}.
\end{equation}

\section{Negative binomial Gibbs sampler updates}\label{app:nb_steps}

\subsection{Sampling $\bbeta_j$}

Let $\omega$ be a \polya-Gamma distributed random variable with parameters $b > 0$ and $c \in \Reals$, denoted $\omega \sim \text{PG}(b, c)$. \cite{polson2013bayesian} proved two useful properties of \polya-Gamma variables. First,
\begin{equation}
\frac{(e^{\psi})^a}{(1 + e^{\psi})^b} = 2^{-b} e^{\kappa \psi} \int_{0}^{\infty} e^{- \omega \psi^2 / 2} p(\omega) \text{d}\omega,
\label{eq:pg_aug_int}
\end{equation}
where $\kappa = a - b/2$ and $p(\omega) = \text{PG}(\omega \mid b, 0)$. And second,
\begin{equation}
p(\omega \mid \psi) \sim \text{PG}(b, \psi).
\end{equation}
Now consider an NB likelihood on $\Y$,
\begin{equation}
    p(\Y \mid \rest)
    = \prod_{n=1}^{N} \prod_{j=1}^{J}
    \frac{(\exp\big\{\bbeta_j^{\top} \varphi(\x_n) \big\})^{y_{nj}}}{(1 + \exp\big\{\bbeta_j^{\top} \varphi(\x_n) \big\})^{y_{nj} + r_j}}.
\end{equation}
Using \eqref{eq:pg_aug_int}, we can express the $nj$-th term in the negative binomial likelihood using the following variable substitutions,
\begin{equation}
    \psi = \bbeta_j^{\top} \varphi(\x_n),
    \quad
    a = y_{nj},
    \quad
    b = y_{nj} + r_j,
    \quad
    \kappa = \frac{y_{nj} - r_j}{2}.
\end{equation}
This gives us
\begin{equation}
\begin{aligned}
&\frac{(\exp\big\{\bbeta_j^{\top} \varphi(\x_n)\big\})^{y_{nj}}}{(1 + \exp\big\{\bbeta_j^{\top} \varphi(\x_n)\big\})^{y_{nj} + r_j}}
\\
&\propto \exp\Big\{ \frac{y_{nj} - r_j}{2} \bbeta_j^{\top} \varphi(\x_n) \Big\} \int_0^{\infty} \exp \Big\{ -\omega_{nj} \frac{(\bbeta_j^{\top} \varphi(\x_n))^2}{2} \big\} p(\omega_{nj}) \text{d} \omega_{nj}
\\
&= \exp\Big\{ -\frac{\omega_{nj}}{2} \Big( \bbeta_j^{\top} \varphi(\x_n) - z_{nj} \Big)^2 \Big\}
\end{aligned}
\end{equation}
where
\begin{equation}
z_{nj} = \frac{y_{nj} - r_j}{2 \omega_{nj}}.
\end{equation}
Finally, note that
\begin{equation}
    \omega \mid \Psi \sim \text{PG}(b, \Psi)
    \implies
    \omega_{nj} \mid \bbeta_j \sim \text{PG}\big(y_{nj} + r_j, \bbeta_j^{\top} \varphi(\x_n) \big).
\end{equation}
If we vectorize across $N$, we can sample each $\bbeta_j$ following \cite{polson2013bayesian}'s proposed Gibbs sampler:
\begin{equation}
\begin{aligned}
    \bbeta_j \mid \bomega_j &\sim \N(\textbf{m}_{\bomega_j}, \textbf{V}_{\bomega_j})
    \\
    \bomega_{j} \mid \bbeta_j &\sim \text{PG}(\y_j + r_j, \bPhi \bbeta_j)
\end{aligned}
\end{equation}
where
\begin{equation}
\begin{aligned}
    \bOmega_j &= \text{diag}([\bomega_{1j}, \dots, \bomega_{Nj}])
    \\
    \V_{\bomega_j} &= (\bPhi^{\top} \bOmega_j \bPhi + \B_0^{-1})^{-1},
    \\
    \textbf{m}_{\bomega_j} &= \textbf{V}_{\bomega_j} (\bPhi^{\top} \bkappa_j + \B_0^{-1} \bbeta_0),
    \\
    \bkappa_j &= (\y_j - r_j) / 2
\end{aligned}
\end{equation}

\subsection{Sampling $r_j$}

Consider the hierarchical model
\begin{equation}
\begin{aligned}
    y_{nj} &\sim \NB(r_j, p_{nj})
    \\
    r_j &\sim \gammadist(a_0, 1/h) 
    \\
    h &\sim \gammadist(b_0, 1/g_0).
\end{aligned}
\end{equation}
\cite{zhou2012augment} showed we can sample $r$ as follows:
\begin{equation}
    r_j \sim \gammadist\Big(L_j, \frac{1}{- \sum_{n=1}^{N} \log(\max(1 - p_{nj}, -\infty))} \Big).
\end{equation}
where
\begin{equation}
    L_j = \sum_{n=1}^{N} \sum_{t=1}^{\ell_j} u_{n\ell},
    \qquad
    u_{n\ell} \sim \log(p_{nj}),
    \qquad
    \ell_j \sim \poisson(-r_j \ln(1-p_{nj})).
\end{equation}

Zhou has released code\footnote{\url{https://mingyuanzhou.github.io/Softwares/LGNB_Regression_v0.zip}}.

\section{Multinomial Gibbs sampler updates}\label{app:multinomial_steps}

\cite{linderman2015dependent} showed that after representing the multinomial distribution as a factorization $J-1$ binomial distributions, we can introduce \polya-gamma random variables to show that the joint distribution of the data $\y_n$ and augmenting variables $\bOmega_n$ is
\begin{equation}
p(\y_n, \bOmega_n)
=
\N\Big(
\bpsi \mid \bOmega_n^{-1} \kappa(\y_n), \bOmega_n^{-1}
\Big)
\end{equation}
where
\begin{equation}
\begin{split}
    \bOmega_n &= \text{diag}([\bomega_{n1}, \dots, \bomega_{n(J-1)}]), \quad \kappa(\y_n) = \y_n - C(\y_n) / 2
    \\
    C(\y_n) &= [C_{n1}, \dots C_{n(J-1)}]^{\top}, \quad C_n = \sum_{j} y_{nj}, \quad C_{nj} = C_n - \sum_{i < j} y_{ni}.
\end{split}
\end{equation}
If we set
\begin{equation}
\bpsi = [\x_n^{\top} \bbeta_1, \dots, \x_n^{\top} \bbeta_{J-1}]^{\top},
\end{equation}
then the marginal for a single $\bbeta_j$ is
\begin{equation}
\N
\Big(
\x_n^{\top} \bbeta_j \mid \frac{\kappa_{nj}}{\omega_{nj}}, \frac{1}{\omega_{nj}}
\Big)
\end{equation}
where $\kappa_{nj} = y_{nj} - \frac{C_{nj}}{2}$. So this gives us a posterior w.r.t. $\bbeta_j$ as
\begin{equation}
p(\bbeta_j \mid \y_j, \X)
\propto
p(\bbeta_j) \prod_{n=1}^{N} \frac{1}{\sqrt{2 \pi} \omega_{nj}} \exp\Big\{ -\frac{\omega_{nj}}{2} \Big(\x_n^{\top} \bbeta_j - \frac{\kappa_{nj}}{\omega_{nj}} \Big)^2 \Big\}
\end{equation}
We can vectorize this across $N$ as
\begin{equation}
\begin{aligned}
&p(\bbeta_j) \prod_{n=1}^{N} \frac{1}{\sqrt{2 \pi} \omega_{nj}} \exp\Big\{ -\frac{\omega_{nj}}{2} \Big(\x_n^{\top} \bbeta_j - \frac{\kappa_{nj}}{\omega_{nj}} \Big)^2 \Big\}
\\
&= p(\bbeta_j) \frac{1}{\sqrt{2\pi}} \Big[ \prod_{n=1}^{N} \frac{1}{\omega_{nj}} \Big] \exp\Big\{ \sum_{n_1}^{N} \Big[ -\frac{\omega_{nj}}{2} \Big(\x_n^{\top} \bbeta_j - \frac{\kappa_{nj}}{\omega_{nj}} \Big)^2 \Big] \Big\}
\\
&\propto p(\bbeta_j) \exp\Big\{ -\frac{1}{2} 
(\z_j - \mathbf{X} \bbeta_j)^{\top} \bOmega_j
(\z_j - \mathbf{X} \bbeta_j)
\Big\}
\end{aligned} \label{eq:multinomial_posterior}
\end{equation}
where
\begin{equation}
\begin{aligned}
\bOmega_j &\equiv \text{diag}([\omega_{1j}, \dots, \omega_{Nj}])
\\
\mathbf{z}_j &\equiv \begin{bmatrix}
\frac{\kappa_{1j}}{\omega_{1j}} & \dots & \frac{\kappa_{Nj}}{\omega_{Nj}}
\end{bmatrix}^{\top}.
\end{aligned}
\end{equation}
This is the same formulation as in Section 3.1 of \cite{polson2013bayesian}, and we can apply his main result:
\begin{equation}
    \bbeta_j \mid \y_j, \bOmega_j \sim \N(\m_{\bomega_j}, \V_{\bomega_j})
\end{equation}
where
\begin{equation}
\begin{aligned}
\V_{\bomega_j} &= (\mathbf{B}_0^{-1} + \mathbf{X}^{\top}\bOmega_j \mathbf{X})^{-1}
\\
\m_{\bomega_j} &= \mathbf{B}_0^{-1} \bbeta_j + \mathbf{X}^{\top} \boldsymbol{\kappa}_j.
\end{aligned}
\end{equation}





